\documentclass{article}

\usepackage{microtype}
\usepackage{graphicx}
\usepackage{subfigure}
\usepackage{booktabs} 
\usepackage{amsmath}
\usepackage{amssymb}
\usepackage{xcolor}

\usepackage{hyperref}

\newcommand{\R}{\mathbb{R}}

\usepackage[accepted]{icml2021}

\icmltitlerunning{Data Considerations in Graph Representation Learning for Supply Chain Networks}

\begin{document}

%

\twocolumn[\icmltitle{Data Considerations in Graph Representation Learning for Supply Chain Networks}
\begin{icmlauthorlist}
\icmlauthor{Ajmal Aziz}{cam_engineering}
\icmlauthor{Edward Elson Kosasih}{cam_engineering}
\icmlauthor{Ryan-Rhys Griffiths}{cam_phys}
\icmlauthor{Alexandra Brintrup}{cam_engineering}
\end{icmlauthorlist}

\icmlaffiliation{cam_engineering}{Department of Engineering, University of Cambridge}
\icmlaffiliation{cam_phys}{Department of Physics, University of Cambridge}

\icmlcorrespondingauthor{Edward Elson Kosasih}{eek31@cam.ac.uk}

\icmlkeywords{Graph Representation Learning, Machine Learning, Supply Chains, Knowledge Graphs, Inductive Knowledge Graph Completion, ICML}
\vskip 0.3in]






\printAffiliationsAndNotice{}  

\begin{abstract}
Supply chain network data is a valuable asset for businesses wishing to understand their ethical profile, security of supply, and efficiency. Possession of a dataset alone however is not a sufficient enabler of actionable decisions due to incomplete information. In this paper, we present a graph representation learning approach to uncover hidden dependency links that focal companies may not be aware of. To the best of our knowledge, our work is the first to represent a supply chain as a heterogeneous knowledge graph with learnable embeddings. We demonstrate that our representation facilitates state-of-the-art performance on link prediction of a global automotive supply chain network using a relational graph convolutional network. It is anticipated that our method will be directly applicable to businesses wishing to sever links with nefarious entities and mitigate risk of supply failure. More abstractly, it is anticipated that our method will be useful to inform representation learning of supply chain networks for downstream tasks beyond link prediction.\footnote{Code available at: \hyperlink{https://anonymous.4open.science/r/Link-Prediction-Supply-Chains-3D76/README.md}{https://anonymous.4open.science/r/Link-Prediction-Supply-Chains-3D76/README.md}}


\end{abstract}


\section{Introduction}

\begin{figure}[htbp!] 
\centering    
\includegraphics[width=.45\textwidth]{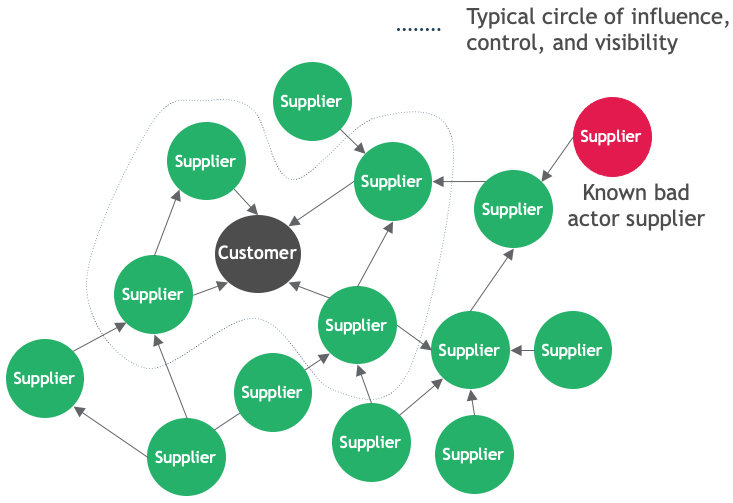}
\caption[KG]{Toy supply chain network demonstrating how geographic, structural, and reputational risk can manifest in extended supply chain networks.}
\label{fig:supply_network}
\end{figure}

Manufacturing firms with non-trivial product offerings scale up by procuring subcomponents, services, or capabilities  \cite{barney1991}. Inevitably, due to labour cost arbitrage and an ever increasing focus on cost efficiencies, supply networks have become more global as firms position themselves to optimise profitability. Whilst globalisation and outsourcing can have financial benefits and lead to faster time to market for manufactured goods, a supply network leads to structural dependencies amongst firms and subsequent concentration of risk, leaving value chains vulnerable to disruptions. The effects of globalisation mean that individual firms have little control or visibility over their extended supply network, exacerbating the risk of disruption. 

In particular, the lack of visibility may result in firms procuring goods and services from firms which are known to perform nefarious activities, examples of which include, but are not limited to, the engagement of child labour, unsustainable business practices, and more general violations of employment law. An illustrative example of the structure of a complex supply network is given in \autoref{fig:supply_network}, where the focal firm (customer) remains unaware of a Tier 2 supplier and is also supplied by a Tier 3 supplier.


Recently, methods that leverage web scraping, entity recognition, and labelling have been proposed to provide transparency of the supply chain \cite{WichmannBBWM20}. In these method, entity recognition is used to derive nodes with edges being built through binary classification applied to text data on the entities. There are two main drawbacks to Natural Language Processing (NLP) based approaches: (i) it is implicitly assumed that all procurement activities are published as articles or metadata on the internet,  and (ii) they are not statistically or otherwise verifiable.


In this work, we propose an automated approach to synthesise an appropriate representation for a downstream link prediction task. It is viewed that automated approaches may complement methods that gather incomplete information and help towards statistical verification of links that have been found. Specifically, we:


\begin{enumerate}
    \item Introduce the first method to learn a heterogeneous graph (knowledge graph) of supply chain network data.
    \item Leverage the learned representation to achieve state-of-the-art performance on link prediction using a relational graph convolution network.
\end{enumerate}


\section{Background}

\subsection{Supply Chain Networks as Graphs}
Representing supply chain networks as graphs was first proposed by~\cite{choi_supply_2001}. Since then, researchers have studied the impacts of ripple effects \cite{chauhan2020relationship}, \cite{dolgui2018ripple}, demonstrated that supply chain networks naturally form hubs and exhibit scale-free characteristics , and even trained algorithms to locate hidden links in these networks~\cite{Brintrup2018} using manually-specified homogeneous graphs (single edge type and node type cf. Section 4). In this work, we build on this body of work by developing a heterogeneous supply chain graph representation that yields improved performance in the downstream task of link prediction.

\begin{figure}[htbp!] 
\centering
\includegraphics[width=.48\textwidth]{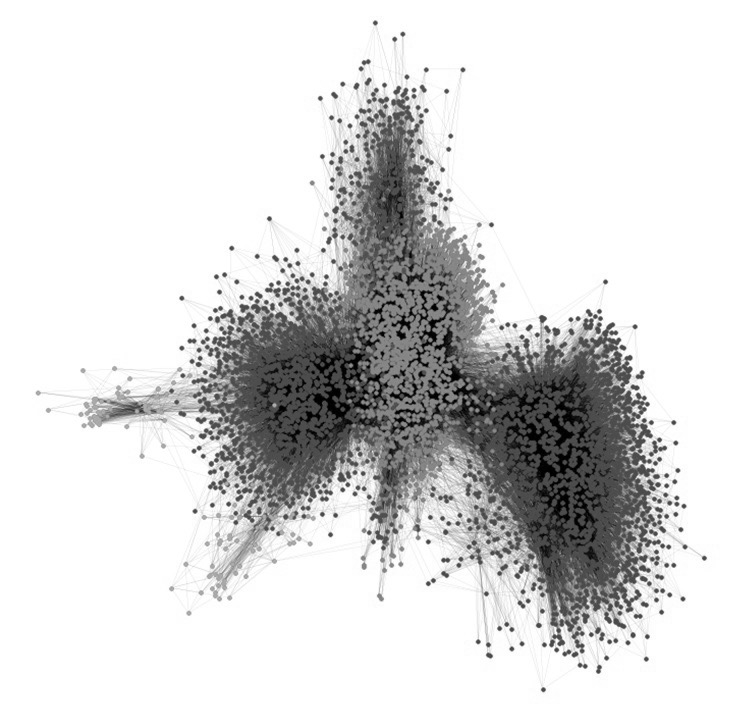}
\caption[KG]{A Depiction of the global automotive supply chain network as a homogeneous graph.}
\label{fig:kg_extract}
\end{figure}


\subsection{Supply Chain Link Prediction}


\begin{figure*}[htbp!] 
\centering    
\includegraphics[width=.8\textwidth]{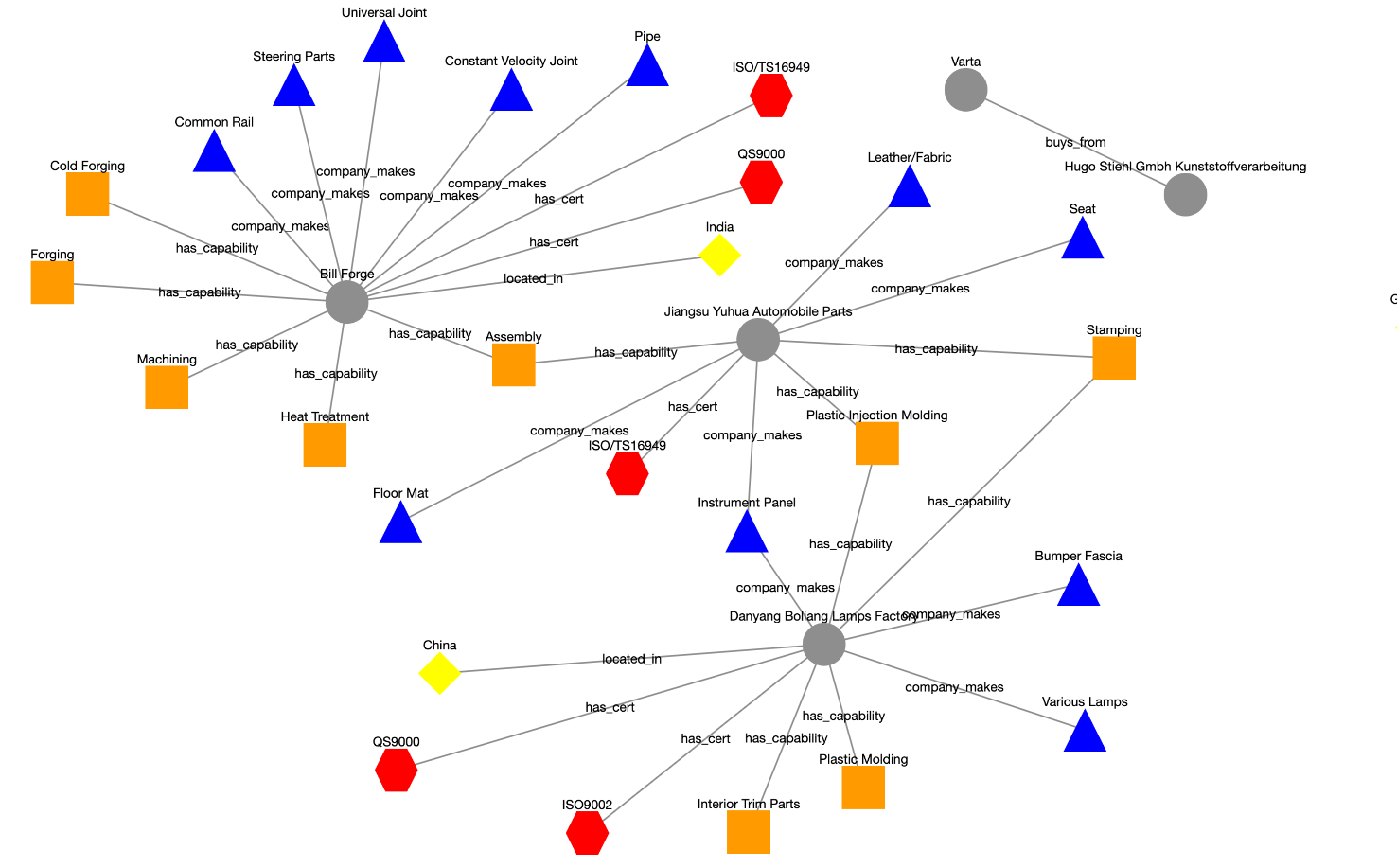}
\caption{An illustrative subgraph from the developed supply chain knowledge graph, composed of multi-type nodes and edges.}
\label{fig:kg_extract_full}
\end{figure*}

\textbf{Link Prediction:} Various techniques have been proposed for link prediction in domains beyond supply chain applications. One of the most commonly-used techniques is based on computing similarity between pairs of nodes. Such similarities are derived based on handcrafted heuristics such as node degrees, or the number of common neighbours; these include the Jaccard Coefficient \citep{libennowell_link-prediction_2007}, Katz \citep{katz_new_1953}, LHN Index \citep{leicht_vertex_2006}, Preferential Attachment \citep{barabasi_emergence_1999}, Adamic-Adar \citep{adamic_friends_2003}, Resource Allocation \citep{zhou_predicting_2009} and path-based similarity \citep{lu_similarity_2009}. 


While many existing heuristics-based similarity techniques could work well in practice, they rely on domain experts to handcraft features. Given that we work with larger datasets with more attributes, manually defining such formulae is expensive. Additionally, while handcrafted heuristics can work well in a particular application, transferring them to different contexts is likely to fail. For instance, \citep{kovacs_network-based_2019} shows that Common Neighbours (CN), a heuristic used in social network analysis, fails to perform in protein graph networks. This is due to the inductive bias stemming from CN's assumption of homophily (similar nodes are connected), an assumption that does not hold in protein networks. Such issues have also been observed in supply chains \citep{brintrup_predicting_2018}. 

Attempts to extract features automatically have been made by node embedding algorithms such as DeepWalk \citep{perozzi_deepwalk_2014}, LINE \citep{tang_line_2015} and node2vec \citep{grover_node2vec_2016}. Here, nodes are represented as vectors derived from topological features obtained by performing various forms of random walk within the neighborhood of the nodes. Link prediction then becomes a binary classification task, where a decoder scores a pair of node embeddings to calculate if there is a high likelihoood of an edge forming between them.

Recent approaches to extract more complex node embeddings are obtained by using graph neural networks (GNNs) (\citep{hamilton_graph_2020}, \citep{bruna_spectral_2014}, \citep{duvenaud_convolutional_2015}, \citep{kipf_semi-supervised_2017}, and \citep{niepert_learning_2016}). GNNs have outperformed many existing algorithms across various domains such as airline carrier networks, citation networks, political blogs, protein interactions, power grids, router-level internet and E. coli metabolite reactions  (\citep{zhang_weisfeiler-lehman_2017}, \citep{zhang_link_2018}, \citep{zhang_revisiting_2020}, \cite{huang_graph_2021}, \cite{teru_inductive_2020}). 

While GNNs have been applied to extract node embeddings, they may also be used to learn representations of a triplet (a pair of nodes with an edge between them). One implementation of such a GNN is called the relational graph convolutional network.

\textbf{Relational Graph Convolutional Networks (RGCNs):} generate latent representations for entities within multi-relational graphs (or knowledge graphs) for downstream graph reasoning tasks. \cite{1schlinchtkrull2018}. Our approach begins by leveraging the GraphSAGE architecture proposed by \citet{2017_Hamilton} to learn functions that  inductively generate node embeddings for all entities in the knowledge graph. The inductive learning paradigm is chosen because supply chain networks evolve over time as companies (which act as autonomous agents in our network representation) choose their locations, product offerings, or procurement relationships. All entity types within the knowledge graph are initialised with a random embedding vector. The set of features for all nodes $\mathbf{X} \in \R^{|\mathcal{V}|\times d}$ are chosen at random where $d$ is the dimensionality of the feature vector associated with the nodes and is treated as a hyperparameter to be tuned during cross validation.


\section{Our Approach}

\subsection{Learning a Heterogeneous Graph Representation of a Supply Chain Network}


The formal definition of a knowledge graph varies between application fields. For the purposes of graph representation learning over supply chain networks, a definition in line with ~\citet{palumbo2020} is adopted. In this paradigm, a knowledge graph can be conceptualised as a 3-tuple/triplet $K = (\mathcal{V}, \mathcal{E}, \mathcal{O})$ where $\mathcal{V}$ is the set of entities (or nodes), $\mathcal{E} \subseteq \mathcal{V} \times \mathcal{V}$ is the set of relations, and $\mathcal{O}$ is the ontology of the knowledge graph.

\begin{figure}[htbp!] 
\centering    
\includegraphics[width=.48\textwidth]{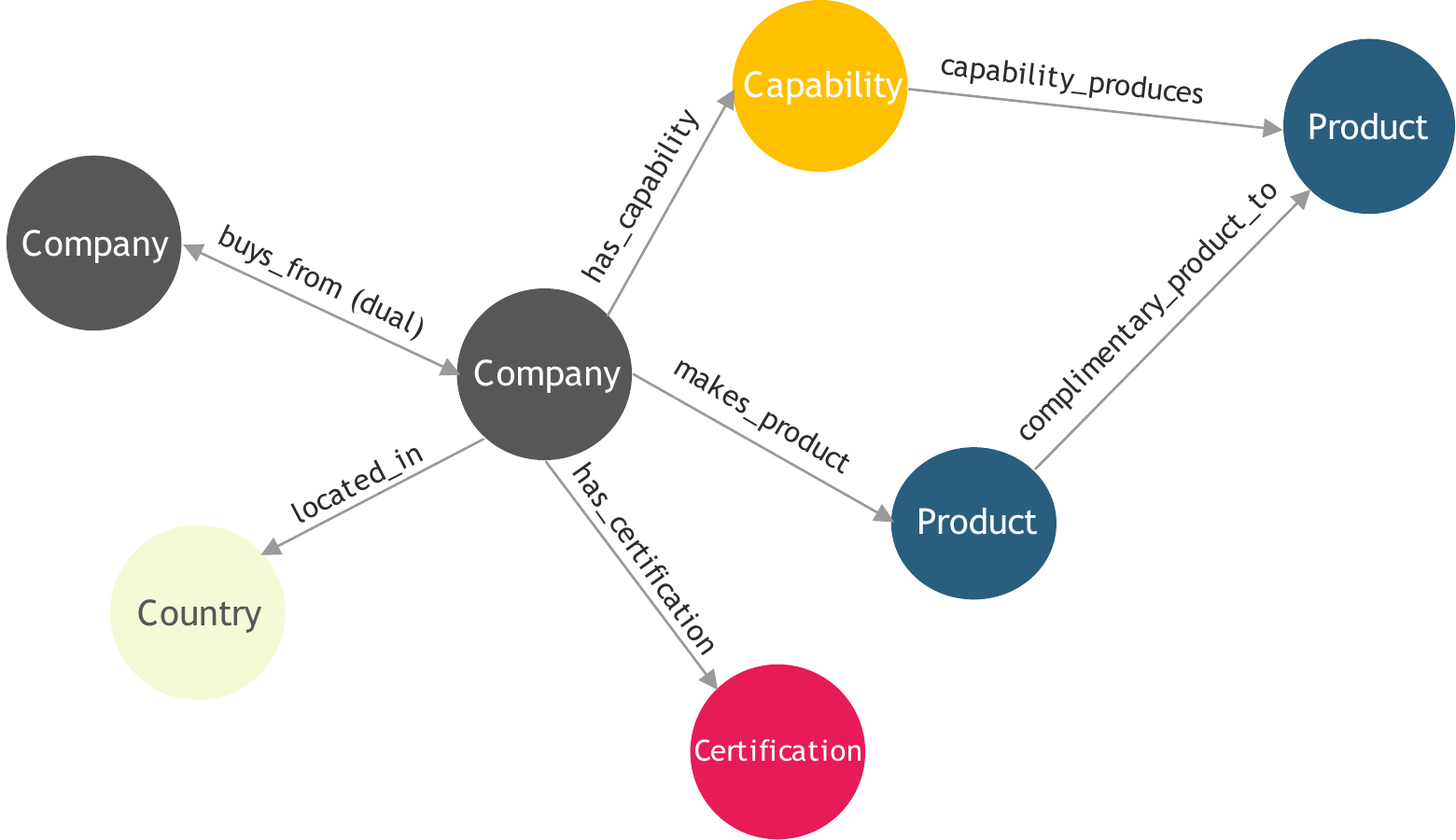}
\caption{Developed ontology to populate the supply chain knowledge graph.}
\label{fig:ontology_kg}
\end{figure}

\textbf{The ontology:} defines the set of entity types, $\Lambda$, and the set of relation types $\mathcal{R}$. Additionally, it assigns nodes to their entity type, $\mathcal{O}: u \in \mathcal{V} \mapsto \Lambda$, and entity types to their related properties, $\mathcal{O}: \epsilon \in \Lambda \mapsto \mathcal{R}_{\epsilon} \in \mathcal{R}$. Effectively, the ontology defines the underlying data structure of a knowledge graph. Within this context, the set of entity types comprises $\{$\verb|Company|, \verb|Capability|, \verb|Certification|, \verb|Product|, and \verb|Country|$\}$ where $|\Lambda| = 5$ . The corresponding set of edges between entities ($\mathcal{R}_{\epsilon} \in \mathcal{R}$) is defined with business-specific use cases in mind. A pictorial representation of the defined ontology is shown in \autoref{fig:ontology_kg}. Considering \autoref{fig:ontology_kg} for entity type \verb|Country|, only a single relation type, \verb|located_in|, is allowed for triplets containing the entity type \verb|Country|.

\begin{figure}[htbp!] 
\centering    
\includegraphics[width=.49\textwidth]{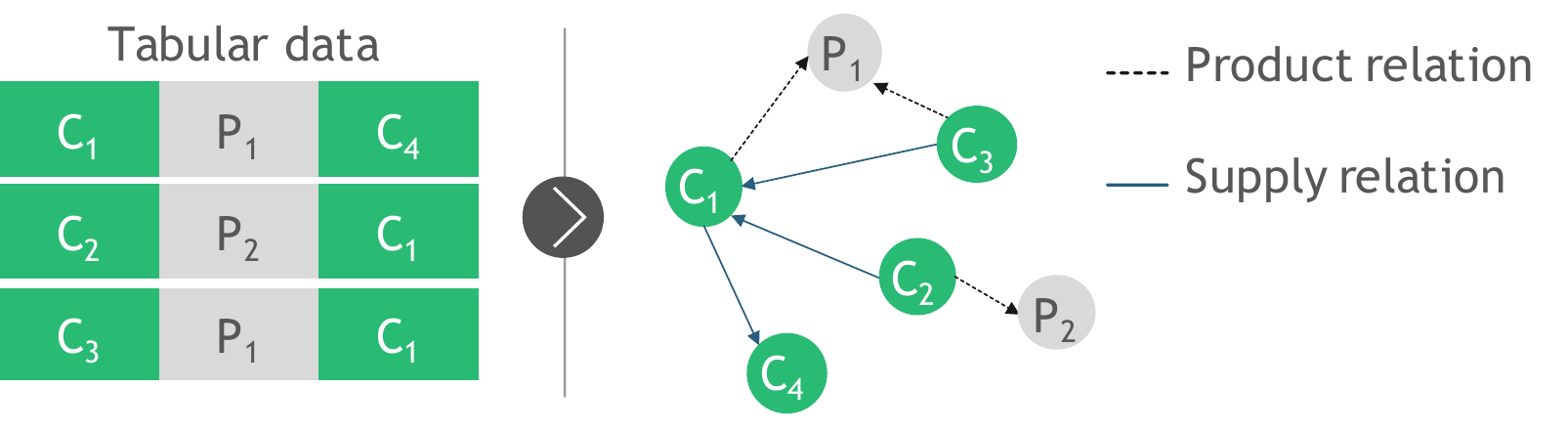}
\caption{Example of how tabular data is converted to a knowledge graph for (only two relation types ($C_n$ for companies, and $P_n$ for products)) according to our defined ontology.}
\label{fig:table_converted_to_graph}
\end{figure}

\textbf{Populating the knowledge graph}: The ontology is populated through a tabular data structure comprising (incomplete) attribute information about companies within the automotive sector\footnote{The data is obtained from MarkLines, a company that specialises in automotive supply chain data collection.}. The tabular data is converted into multiple multipartite graphs to derive relations. For an indicative example, \autoref{fig:table_converted_to_graph} demonstrates this procedure for two relation types: (company, \verb|buys_from|, company) and (company, \verb|makes_product|, product). Where relationships could not be deduced from the tabular data, bipartite projections were taken over the entity set where information was missing. This is a crucial step as complementary capability and product offerings may embed inductive bias when predicting \verb|buys_from| relations. 

Some links included within the ontology were not immediately available in collected data but could be deduced. In this case, a co-occurrence frequency was used to derive these relations. The intuition here is that if a company possesses a capability (e.g. Plastic Injection Moulding) and produces products (Seat Belts, Bumpers, etc.), then enough instances of co-occurrence of capabilities with the same product would imply that the capability and product can be tied into the \verb|capability_produces| relation. 

\begin{figure}[htbp!] 
\centering    
\includegraphics[width=.48\textwidth]{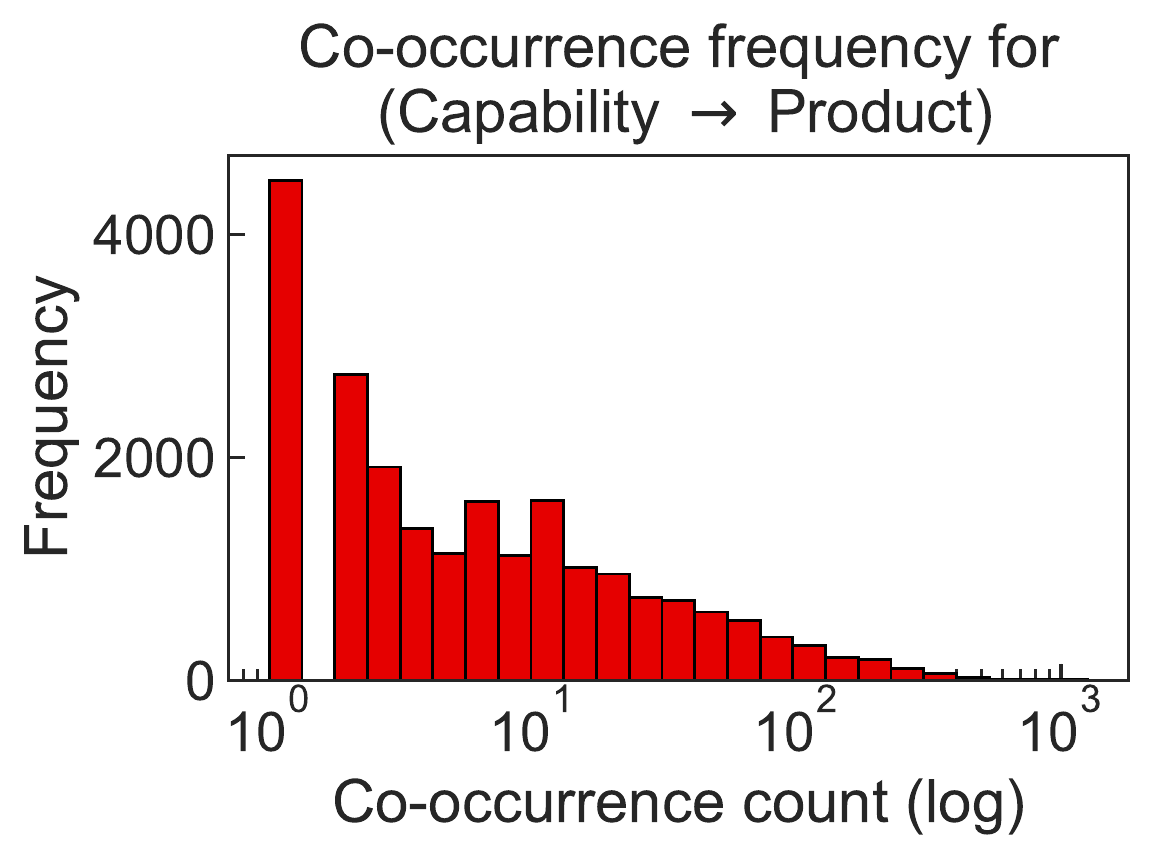}
\caption{Capabilities and products co-occurrence frequency weights.}
\label{fig:weights_capability_product_projection}
\end{figure}

The histogram of co-occurrence frequency is shown in \autoref{fig:weights_capability_product_projection}. As the data exhibits noise potentially due to spurious information, a cutoff threshold is required to filter relations based on co-occurrence frequency. This threshold is treated as a hyperparameter during training and can be optimised for whichever edge type a company may deem the riskiest. For example, if a company is interested in geographic risk, then the cutoff threshold is optimised for predicting \verb|buys_from| relationships successfully.

\begin{figure}[htbp!] 
\centering    
\includegraphics[width=.48\textwidth]{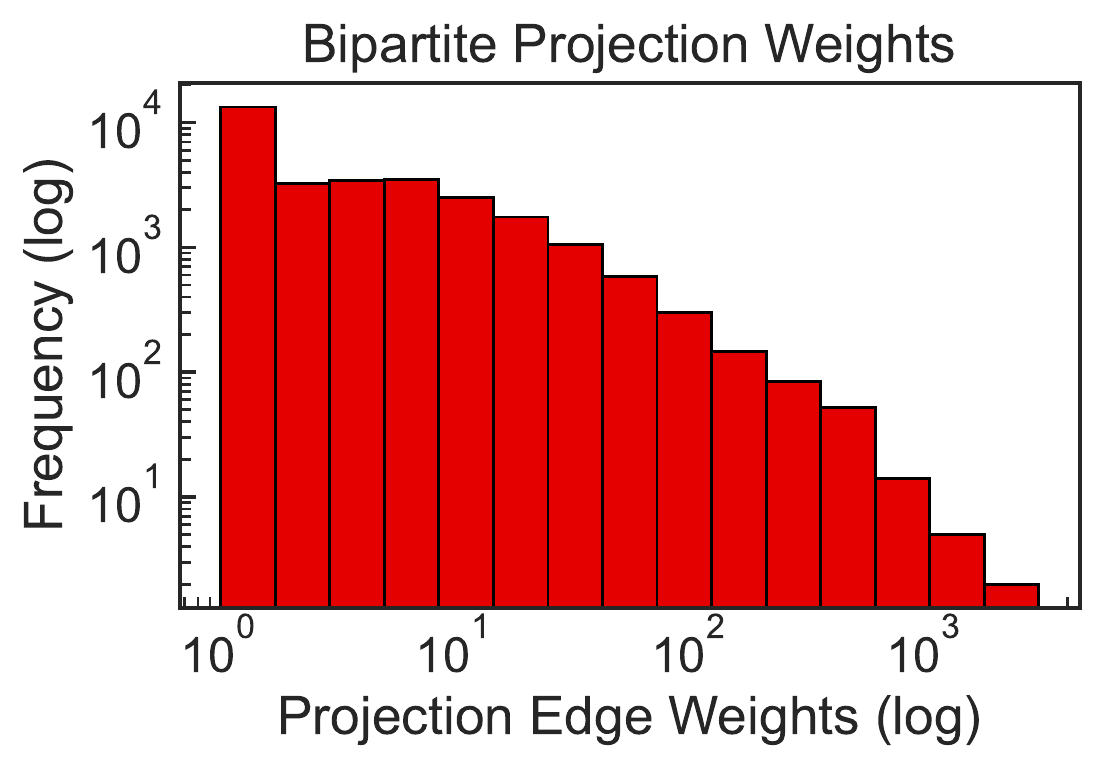}
\caption{Bipartite projection weights for product entities. $\mathcal{V}_A$ and $\mathcal{V}_B$ denote the disjoint .}
\label{fig:bipartite_projection}
\end{figure}

The other edge type that has to be deduced from data is $\tau = \verb|complimentary_product_to|$. For this edge type, a bipartite graph consisting of relation type $\tau=\verb|makes_product|$ between companies and their respective product portfolio is leveraged. A bipartite projection is taken onto the product entities where the weights in the projection space indicate the number of times companies purchased similar products. ~\autoref{fig:bipartite_projection} shows the distribution of edge weights in the projection space. The cutoff threshold for introducing \verb|complimentary_product_to| relations is also treated as a hyperparameter during training. 

\textbf{Triplets} or labelled directed edges are represented as factual tuples $(u, \tau, v)$ for $u, v \in \mathcal{V}$ and $\tau \in \mathcal{R}$. For example, we have edge types $\tau = \verb|has_capability|$ and the edge $(u, \verb|has_capability| , v)$ where $u=\verb|Bill Forge|$ and $v=\verb|Forging|$ indicate known information about a company and its capability since the relation type is restricted between entity types \verb|Company| and \verb|Capability|. 
~\autoref{tab:entities} and ~\autoref{tab:triplets} convey the extracted entities totalling $\sim$161k and extracted facts totalling $\sim$647k respectively. 

Finally, the learning objective is the same as knowledge graph completion, and is geared towards predicting missing edges to complete the knowledge graph representation.

\begin{table}[htbp!]
\centering
\begin{tabular}{lc}
\toprule
\textbf{Entity Type}                         & \textbf{Count}  \\ \midrule
company (e.g. General Motors) & 41,826 \\
product (e.g. Floor mat)      & 119,618 \\
country (e.g. Germany).       & 74        \\
capability (e.g. Machining)   & 36      \\
certification (e.g. ISO9001)  & 9       \\ \midrule
Total                         & 161,563 \\ \bottomrule
\end{tabular}
\caption{Entity count contained within the supply chain knowledge graph.}
\label{tab:entities}
\end{table}

\begin{table}[htbp!]
\centering
\begin{tabular}{lc}
\toprule
\textbf{Triplet Type}                                  & \textbf{Count}   \\ \midrule
(capability, capability\_produces, product) & 21,857  \\
(company, buys\_from, company)                                   & 88,997  \\
(company, has\_capability, capability)                & 83,787  \\
(company, has\_cert, certification)                               & 32,654  \\
(company, located\_in, country)                             & 40,421  \\
(company, makes\_product, product)                                  & 119,618 \\
(product, complimentary\_product\_to, product)                  & 260,658 \\ \midrule
Total                                       & 647,992 \\ \bottomrule
\end{tabular}
\caption{Triplet count contained within the supply chain knowledge graph.}
\label{tab:triplets}
\end{table}

\subsection{Loss Function for Link Prediction}
The latent embeddings $\mathbf{h}_u$ for nodes $u \in \mathcal{V}$ are generated using the GraphSAGE architecture in the minibatch setting. In the GraphSAGE paradigm, trainable functions are learned to generate compact embeddings by sampling and aggregating features from local neighbourhoods of nodes to be used in downstream tasks (link prediction in our case). The aggregator function $\textsc{aggregate}^\tau_k$  $\forall k \in \{1, \ldots, K\}$ for depth $K$ and $\tau \in \mathcal{R}$ as well as trainable weight matrices used in updating latent embedding $\mathbf{W}^K_\tau$ for $\tau \in \mathcal{R}$ are trained to minimise the binary cross entropy loss across all relation types. This choice of link prediction loss is similar to that proposed by~\citet{2018_Schlichtkrull} and is given as: 
\begin{equation}
\begin{split}
    \mathcal{L} = \sum_{(u,\tau, v), y} &y \text{ log} f(u, \tau, v) + \\&(1-y)\text{ log}\left( 1-f(u, \tau, v) \right)
\end{split}
\end{equation}

Where triples $(u, \tau, v)$ for $(u, v) \in \mathcal{V}$ with relation $\tau \in \mathcal{R}$ are scored according to $f(u, \tau, v)$ based on an indicator $y \in \{0, 1\}$ denoting whether or not the triplet exists (detailed further in~\autoref{sec:experiments}). The score $f(.)$ is derived based on the $K$-th node embeddings for source and destination nodes, $\mathbf{h}^K_u$ and $\mathbf{h}^K_v$ respectively, and was chosen as $f(u, \tau, v) = (\mathbf{h}^K_u)^T R_{\tau} \mathbf{h}^K_u$, which is the DistMult scoring function~\cite{yang2015embedding}. In this context, $R_{\tau} \in \R^{d \times d}$ is a diagonal matrix for every relation $\tau \in \mathcal{R}$ and $d$ is the size of the initialised node embeddings. The loss naturally incentivises the model to associate higher scores to observable triples and lower scores for unobserved triples.

\section{Related Work}

The application of link prediction in supply networks has been scarce. To the best of our knowledge, Supply Network Link Prediction (SNLP) \citep{brintrup_predicting_2018} is the only published work to apply link prediction. SNLP was applied on the same automotive supply chain dataset that is used in our work. The authors represent the supply chain network as a homogeneous graph with one type of edge/relation: \verb|buys_from|, unlike our heterogeneous knowledge graph which has multiple relation types. 

This baseline model represents every node with a set of attributes derived from handcrafted heuristics, such as the number of existing suppliers, overlaps between both companies' product portfolios, product outsourcing associations and likelihood of having common buyers. This bears similarity with modern graph node embedding techniques, albeit their representations were not learnable. The approach treats link prediction as a binary classification problem given a pair of nodes with their respective attributes. They report an Area Under the Receiver Operating Curve (AUC) score of 0.76.

\section{Experiments and Results}
\label{sec:experiments}

The task of relational link prediction is to discern whether a given edge $(u, \tau, v)$ is present in $\mathcal{E}^U - \mathcal{E}$ where $\mathcal{E}^U$ is the set of all possible edges and $\mathcal{E}$ is the set of captured edges in the knowledge graph representation. The set $\mathcal{E}^U - \mathcal{E}$ is the set of edges that have not been captured when building the supply chain knowledge graph, or are edges that will present themselves in the future (a new partnership between two companies is formed, new capabilities are invested in, etc.). The learning regime involves cross validation (70\% training, 20\% validation, and 10\% testing) by splitting the set of all actualised triples into a training, validation, and test set. Negative triplets (triplets which are not facts in the knowledge graph) are then corrupted by either swapping the source or destination nodes ($u$ and $v$) or by uniformly sampling a new relation type between the source and destination nodes. Models are assessed based on their capability to differentiate between factual and non-factual triplets. The task is therefore distilled into a binary classification task (for all relation types), and the commonly-used Area Under the Receiver Operating Curve (AUC) is used to assess model performance. To the best of our knowledge, the best reported AUC for this task is 0.76 (for the \verb|buys_from| relation in our context). As shown in~\autoref{tab:experimental_results}, our multi-relational model outperforms the existing baseline and extends the prediction task to multiple relations.

\begin{table}[htbp!]
\begin{tabular}{@{}lccc@{}}
\toprule
Relation Type              & Train & Validation & Test \\ \midrule
makes\_product             & 1.000            &      0.996 &  0.989\\
has\_cert                  & 0.825        &     0.591 &  0.430 \\
complimentary\_product\_to & 0.997  &     1.000           &     1.000        \\
located\_in                & 0.955  &     0.977           &     0.613        \\
has\_capability            & 0.802   &         0.676       &    0.564         \\
buys\_from                 & \textbf{0.932}        &    \textbf{0.862}            &   \textbf{0.877}          \\
capability\_produces       & 0.993      &   1.000        &     1.000        \\ \bottomrule
\end{tabular}
\caption{AUC scores for training, validation, and test graphs. Note, test set results were not recalculated based on retraining with both training and validation edges. Results with bold face outperform existing benchmarks (SNLP). Other table entries represent novel relation types that have not been considered in prior work.}
\label{tab:experimental_results}
\end{table}

\section{Conclusion}

Due to the effects of globalisation, supply chains are becoming more complex, and obtaining visibility into interdependencies within the network has become a tremendous challenge. While better information extraction techniques have been developed, there remains a large gap towards obtaining a complete representation of the network. The raw data alone often has missing information due to a company's propensity to engage in secretive and competitive behavior. This information, however, is particularly important for supply chain practitioners to detect operational risks, such as unfair manufacturing practices and overreliance on certain sole suppliers. Graph representation learning, in the form of link prediction, can help impute such missing data. 

Our paper proposes a novel method for learning a representation of a supply chain network as a heterogeneous graph, allowing us to predict the existence of various type of dependencies, as opposed to the incumbent SOTA (SNLP) approach of predicting just one type of dependency using a homogeneous graph. Moreover, our embeddings are learnable, which may also be responsible for the improved performance relative to SNLP. 

In future work we wish to perform an ablation study to isolate the contributions of the learnable embedding and heterogeneous graph components. An extension of this work will include exploration of graph learning techniques for multi-hop reasoning to detect more complex dependencies associated with paths in the graphs, as opposed to single links.


\bibliography{ref}
\bibliographystyle{icml2021}

\end{document}